\documentclass[letterpaper, 10 pt, conference]{ieeeconf}  % Comment this line out if you need a4paper

\IEEEoverridecommandlockouts                              % This command is only needed if 
\usepackage{graphics}
\usepackage{multirow}
\usepackage{graphics} % for pdf, bitmapped graphics files
\usepackage{epsfig} % for postscript graphics files
\usepackage{times} % assumes new font selection scheme installed
\usepackage{amsmath} % assumes amsmath package installed
\usepackage{amssymb}  % assumes amsmath package installed

\usepackage{array}
\usepackage{booktabs}

\usepackage{hyperref}
\usepackage{caption}
\usepackage[T1]{fontenc}  % To make copy of underscore work

\usepackage{xcolor,pifont}
\usepackage{color}
\usepackage{acronym}

\newcommand*\colourcheck[1]{%
  \expandafter\newcommand\csname #1check\endcsname{\textcolor{#1}{\ding{52}}}%
}
\newcommand*\colourcross[1]{%
  \expandafter\newcommand\csname #1cross\endcsname{\textcolor{#1}{\ding{55}}}%
}
\colourcheck{green}

\definecolor{somegray}{rgb}{0.5, 0.5, 0.5}
\newcommand{\darkgrayed}[1]{\textcolor{somegray}{#1}}
\makeatletter
\newcommand*\titleheader[1]{\gdef\@titleheader{#1}}
\AtBeginDocument{%
  \let\st@red@title\@title
  \def\@title{%
    \vskip-3em
    \bgroup\normalfont\large\centering\@titleheader\par\egroup
    \vskip1.5em\st@red@title}
}
\makeatother

\titleheader{\darkgrayed{This paper has been accepted for publication at the\\
IEEE International Conference on Robotics and Automation (ICRA), London, 2023.
\copyright IEEE}}

\makeatletter
\let\@oldmaketitle\@maketitle% Store \@maketitle
\renewcommand{\@maketitle}{\vspace{30pt}\@oldmaketitle% Update \@maketitle to insert...

\centering
\includegraphics[width=1.0\textwidth]{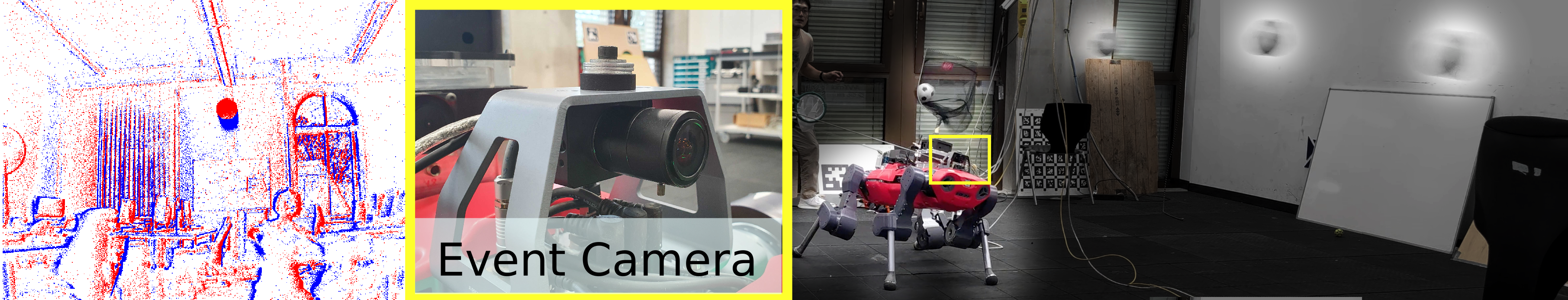}
\captionof{figure}{Event-based agile object catching with a quadrupedal robot, using only onboard perception and control. By using the low latency of events, our robot is able to catch objects at speeds up to 15 m/s with a 83\% success rate.}\label{fig:eyecatcher}%\bigskip%
\vspace{-10pt}
}% ... an image
\makeatother

\title{\LARGE \bf
Event-based Agile Object Catching with a Quadrupedal Robot
}

\author{Benedek Forrai$^{*1}$, Takahiro Miki$^{*1}$, Daniel Gehrig$^{*2}$, Marco Hutter$^1$, Davide Scaramuzza$^2$% <-this % stops a space
 \thanks{
    $^*$ Denotes equal contribution.
    $^1$Benedek Forrai, Takahiro Miki and Marco Hutter are with the Robotic Systems Lab, Department of Mechanical Engineering, ETH Zurich, Switzerland. $^2$Daniel Gehrig and Davide Scaramuzza are with the Robotics and Perception Group, University of Zurich Switzerland (\protect\url{http://rpg.ifi.uzh.ch}). 
    This work was supported by the Swiss National Science Foundation (SNSF) through the National Centre of Competence in Research (NCCR) Robotics, and the European Research Council (ERC) under grant agreement No. 864042 (AGILEFLIGHT).
    }
    }% <-this % stops a space
\begin{document}

\newacro{RL}{Reinforcement Learning}
\newacro{GPU}{Graphics Processing Unit}
\newacro{PPO}{Proximal Policy Optimization}
\newacro{MDP}{Markov Decision Process}

\maketitle
\thispagestyle{empty}
\pagestyle{empty}

%%%%%%%%%%%%%%%%%%%%%%%%%%%%%%%%%%%%%%%%%%%%%%%%%%%%%%%%%%%%%%%%%%%%%%%%%%%%%%%%
\begin{abstract}
% opening statement motivating the use of quadrupedal robots
Quadrupedal robots are conquering various applications in indoor and outdoor environments due to their capability to navigate challenging uneven terrains. 
Exteroceptive information greatly enhances this capability since perceiving
their surroundings allows them to adapt their controller and thus achieve higher levels of robustness.
%A key driving factor of this trend is the use of highly maneuverable, proprioceptive legs that are able to perform a variety of gaits that can be adapted quickly.
% However, proprioception alone does not provide sufficient information to detect or react to falling or flying objects quickly and precisely. 
However, sensors such as LiDARs and RGB cameras do not provide sufficient information to quickly and precisely react in a highly dynamic environment since they suffer from a bandwidth-latency tradeoff. They require significant bandwidth at high frame rates while featuring significant perceptual latency at lower frame rates, thereby limiting their versatility on resource constrained platforms. 
% While LiDARs and RGB cameras can fill this gap by providing additional exteroceptive queues, they typically suffer from a bandwidth-latency tradeoff, requiring signficant bandwidth at high frame rates, while featuring significant perceptual latency at lower frame rates. 
In this work, we tackle this problem by equipping our quadruped with an event camera, which does not suffer from this tradeoff due to its asynchronous and sparse operation. In leveraging the low latency of the events, we push the limits of quadruped agility and demonstrate high-speed ball catching for the first time. We show that our quadruped equipped with an event-camera can catch objects with speeds up to 15 m/s from 4 meters, with a success rate of 83\%. Using a VGA event camera, our method runs at 100 Hz on an NVIDIA Jetson Orin.
\end{abstract}
%%%%%%%%%%%%%%%%%%%%%%%%%%%%%%%%%%%%%%%%%%%%%%%%%%%%%%%%%%%%%%%%%%%%%%%%%%%%%%%%
\section*{Multi-Media Material}
For visual results, see our video at \url{https://youtu.be/FpsVB8EO54M}. Also check out open-source code at \url{https://github.com/uzh-rpg/event-based_object_catching_anymal}.
\section{Introduction}
\begin{figure*}[t]
    \centering
    \includegraphics[width=\linewidth]{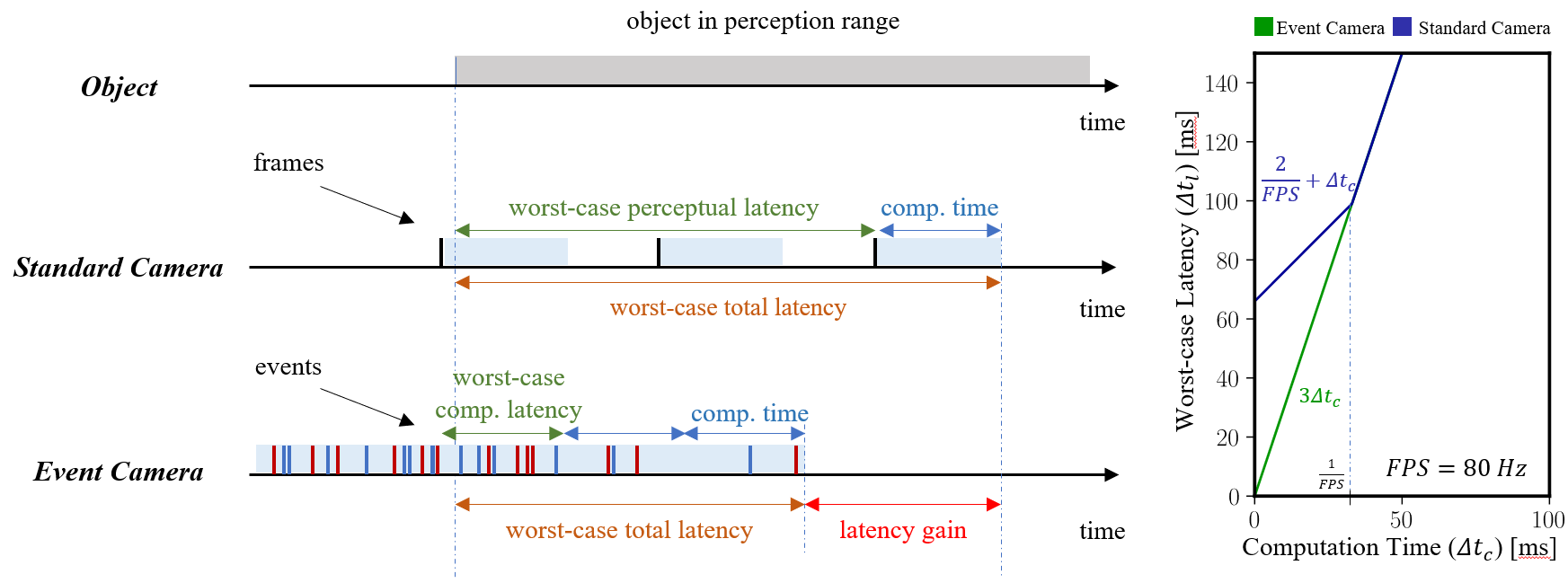}
    \caption{Worst-case total latency comparison of standard camera and event
    camera-based algorithms for object catching. Both algorithms rely on $N\geq 2$ detection of 3D position to fit a ballistic parabola, after which they can infer the impact range for catching.}
    \label{fig:latency_analysis}
    \vspace{-2ex}
\end{figure*}
% motivation: imitate humans in their ability to navigate complex terrains
Humans and animals can perform highly agile maneuvers, which require a combination of fast and precise perception and robust and accurate control. For years roboticists have been inspired by nature and aimed at emulating these capabilities in robotic systems. Recently, quadrupedal robots are reaching animal-like performance on challenging locomotion tasks with exteroception --- such as taking a hike in the Alps \cite{Miki22sciRob} --- and are capable of performing complex maneuvers like recovering from a flipped position~\cite{Lee2019-vj}.

In addition to robust control, many animals also have special retinas with which they perceive their environment, enabling them to perform highly complex maneuvers like a dog catching a frisbee in mid-air. Exteroceptive sensors like LiDARs and RGB cameras have enhanced robots' perception capabilities and enabled more smooth and complex behaviors. However, despite these advancements, current quadruped robots are still far from achieving the same level of agility as animals like cats and dogs.  
Indeed, high-speed object catching with onboard sensing remains a significant challenge for quadrupedal robots. %since it requires agile control and high-speed perception.
%animal-like,  
%One direction toward this goal is the development of quadrupedal robots. Inspired by animals, these robots are now capable of tackling a wide range of environments, like  
% show state of the art in perception aware quadruped navigation

\iffalse
Quadrupedal robots equipped with LiDARs or image sensors have made signficant strides toward mimicking similar capabilities, and can now perform autonomous exploration and mapping of previously unknown and potentially dangerous caves \cite{Tranzatto2022Cerberus}, locomote robustly in uneven forests and even perform a hike in the Alps . 
\fi

% motivate the need for perceiving dynamic objects

\iffalse
While these sensors have traditionally been used to reason about static environments for mapping, or adaptive gait adjustment [], reasoning in dynamically changing scenes remains underexplored. In particular, when faced with sudden events like rocks falling in a cave or flying objects, robots need to be able quickly evade or catch objects.   
\fi

% related work on object catching 
Object catching has been studied with standard RGB cameras \cite{Su2016catching}. However, it still suffers from fundamental limitations due to the bandwidth-latency tradeoff. High frame rates are necessary to react quickly, but these incur high bandwidths. By contrast, lowering the frame rate increases perceptual latency and can severely impact the robot's reaction time and success rate.
% introduce our solution, inspire by nature
Animal vision systems do not suffer from this tradeoff since they do not perceive their environment as a sequence of frames. Event cameras are neuromorphic sensors that try to emulate the working principle of these systems, and are therefore also not affected by this tradeoff. They only measure \emph{changes in intensity} which they transmit as asynchronous, binary spikes with micro-second level latency.  They have been successfully used in high-speed object evasion with drones~\cite{Falanga20scirob,Sanket20icra}. However, it remains unclear how policies devised for quadrotors can be applied to different robotic platforms such as quadrupeds. In this work, we fill this gap, by developing a modular approach to high-speed object catching using an event camera on a quadruped platform equipped with a net, seen in Fig. \ref{fig:eyecatcher}. 

Our approach works by first detecting independently moving objects in the event stream, and then fusing these detections to a ballistic parabolic trajectory estimate. This estimate is then processed by a data-driven control policy to perform the catching maneuver. By using events, our approach enjoys a minimal latency, while benefiting from an expressive data-driven policy for catching objects arriving at different angles. Our contributions are the following: 

\begin{itemize}
    \item We present a method for high-speed ball catching from event-based object trajectory estimations. By leveraging the low latency of events we can track objects at up to 100 Hz on an NVIDIA Jetson Orin.  
    \item We are the first to demonstrate successful event camera-based high-speed object catching with a quadrupedal robot equipped with a net. Our policy  catches objects flying at up to 15 m/s from 4 meters with an 83\% success rate.
    \item We extend the latency analysis in \cite{Falanga19ral} to estimate object trajectories from $N$ detections, and the incurred latencies for standard and event camera-based algorithms. 
\end{itemize}

\begin{figure*}
    \centering
    \vspace{10pt}
    \includegraphics[width=0.95\linewidth]{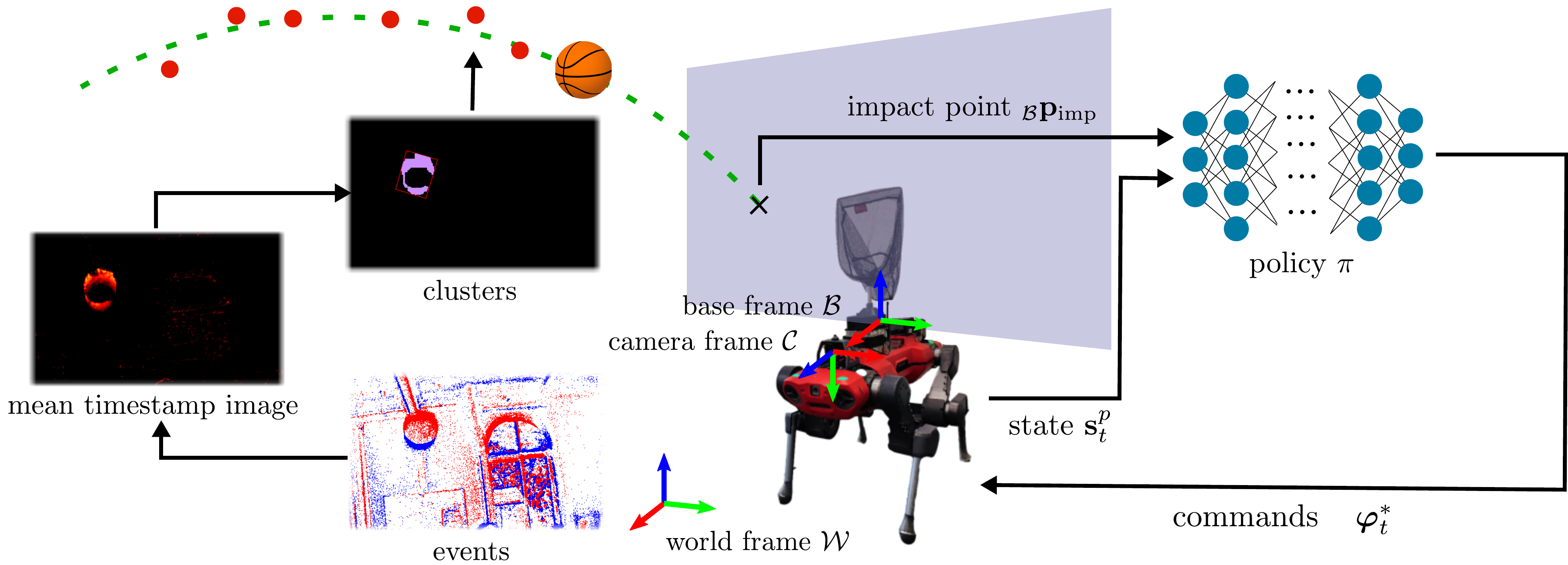}
    \caption{Overview of the approach to ball catching. A quadrupedal robot equipped with an event camera observes raw events of its surrounding. With IMU measurements we compute the motion compensated mean timestamp image from these events, which we segment out moving object clusters. These clusters are projected in 3D and a parabola is fit through them. We compute the impact point of this parabola in a gravity aligned coordinate frame centered at the robot base. The robot policy uses this impact point $_\mathcal{B}\textbf{p}_\text{imp}$ and proprioceptive states $\mathbf{s}^p_t$ to compute target joint deviations $\boldsymbol{\mathbf{\varphi}}^*_t$.}
    \label{fig:approach_overview}
    \vspace{-2ex}
\end{figure*}

\section{Related Work}
\subsection{Dynamic Object Catching in Robotics} 
Early work on dynamic object catching focused on the case where the object trajectory is fully observable and known and has shown successful implementation on a variety of robotics platforms ranging from robotic arms~\cite{Kim14tro} to drones~\cite{Bouffard12icra}. Object trajectories in these cases were detected by a motion capture rig using IR markers. Later work showed successful object catching with onboard perception on a drone \cite{Su2016catching}, however, due to the use of a standard camera, objects speeds were limited to 6 m/s, and the vision task was simplified by mounting bright LED lights on the balls. To date, the work by Falanga et al.~\cite{Falanga19ral} targets the fastest moving objects, flying at up to 10 m/s with a drone. However, it focuses on object evasion, instead of catching, which puts a lower constraint on the robot policy. By contrast, high-speed object catching, the target of this work, requires precise and agile maneuvering of the robot. The work in \cite{Ziyun22ral} is the first to address this task with a linear slider, showing promising results with a learning-based detector. In this work, we take the next step and apply this concept to a complex, legged robot.

\subsection{Advancements in Legged Robotics}
In contrast to the highly maneuverable, lightweight, and easily modeled quadcopters used for onboard catching scenarios in the previous section, legged robots pose a much more sophisticated control problem due to its size and more complex interaction with the environment. 
A typical approach to control this is to use a model-based controller to plan its motion through a physical model and minimize the objective function such as target velocity error. This often requires computational heavy planning, making it challenging to achieve fast and reactive motion~\cite{neuhaus2011comprehensive,jenelten2020perceptive,Grandia2022-vz}.
Recently, a deep reinforcement learning method showed a great capability in complex and agile maneuvers without compromising real-time performance~\cite{Hwangbo2019-jk, lee2020learning, peng2020learning, siekmann2021blindarxiv, Miki22sciRob}.
Utilizing the exteroceptive information was mostly from a separate mapping pipeline which fuses information from the depth sensors~\cite{kim2020vision,fankhauser2018robust,Miki2022-mq}.
However, this approach poses large latency and makes it difficult to handle dynamic environment.
Recently, the attempt of using raw depth images for locomotion task is pursued~\cite{pmlr-v164-yu22a, yang2022learning}, however, achieving a high-speed visual task in a dynamic environment such as catching a flying object with on-board vision is still an open question. Off-board approaches have already shown promising results: a recent work demonstrated a small quadruped capable of acting as a goalie using a hierarchical reinforcement learning approach, choosing different behaviors based on incoming ball trajectories ~\cite{huang2022creating}.
\section{Worst Case Latency Analysis}
As stated in the introduction, event cameras do not suffer from a latency-bandwidth tradeoff, as do RGB cameras. In this section we will clarify this point, by studying the worst-case latency incurred by algorithms $\mathcal{A}$ operating on images or events. We will assume that the goal of $\mathcal{A}$ is to find the parabola describing the ballistic trajectory of the object to be caught, and that this parabola is estimated via a sequence of 3D position and/or 3D velocity measurements. We measure the latency of $\mathcal{A}$ as the worst-case waiting period to collect a minimum number of measurements that describe the parabola. In this work, we focus on \emph{two-shot} methods, that require two consecutive 3D position measurements but discuss the potentially interesting class of \emph{one-shot} methods that measure 3D position and velocity simultaneously and can therefore be even faster. We will discuss the latency considerations for two-shot algorithms designed for standard cameras and event cameras next, summarized in Fig. \ref{fig:latency_analysis}. 

\noindent\textbf{Standard camera}
Algorithms for processing standard images fall in the category of two-shot methods, since single images do not carry motion information. We assume a camera with inter-frame time interval $\Delta t_\text{FPS}=\frac{1}{\text{FPS}}$ and an algorithm with processing time $\Delta t_c$. In the worst case the robot has to wait a period $\max\left\{\Delta t_c, \Delta t_\text{FPS}\right\}$ before it can execute a detection because it has to wait for the next frame to arrive, or a previous detection cycle to terminate (see Fig. \ref{fig:latency_analysis} middle). Then it can initiate two consecutive detections, which together take $\max\left\{\Delta t_c, \Delta t_\text{FPS} \right\}+\Delta t_c$, where the first and second term measure the latency for detecting on the first and second frame respectively. If the computation time is lower than the framerate, the detector still needs to wait until the second frame arrives, so its latency is lower-bounded by $\Delta t_\text{FPS}$. As a result, the algorithm has a worst-case latency 
\begin{align}
     \Delta t_l = \begin{cases}
        2\Delta t_\text{FPS} + \Delta t_c \quad &\text{ if $\Delta t_c <  \Delta t_\text{FPS}$}\\
        3\Delta t_c &\text{ otherwise}.
        \end{cases}
\end{align}
\textbf{Event Camera}
For an event-based algorithm $\Delta t_\text{FPS}\approx 0$, so the initial waiting period before detection can start is dominated by a previous detection cycle (see Fig. \ref{fig:latency_analysis} bottom). In that case, the total worst-case latency becomes 
\begin{align}
    \Delta t_l = 3\Delta t_c
\end{align}
However, since events also carry motion information, an algorithm may try to simultaneously extract 3D position and velocity, and thus this latency can be reduced to $\Delta t_l = 2\Delta t_c$. As a result, an event-based algorithm can have lower latency when the following conditions are satisfied
\begin{align}
    \Delta t^e_c < \Delta t_{\text{FPS}} \quad \text{ for two-shot methods}\\ 
    \Delta t^e_c < \Delta t_{\text{FPS}} + \frac{\Delta t^i_c}{2} \quad \text{ for one-shot methods.} 
\end{align}
In the next section we present a two-shot approach to object catching, and thus our method will benefit more from a fast execution time, outperforming RGB camera-based approaches for similar inter-frame intervals. We leave the study of one-shot approaches for future work.
\section{Approach}

Our object-catching algorithm consists of two main components: A visual frontend, which is tasked with object trajectory estimation based on event-based visual input, and a planning backend that converts the predicted trajectory to low-level motor commands for the quadruped robot. Our full method is illustrated in Fig. \ref{fig:approach_overview} and described further below.

\subsection{Visual Frontend}
Our trajectory estimator is inspired by the moving object detector used in \cite{Falanga20scirob}, followed by trajectory fitting and filtering steps. For completeness, we will briefly summarize the method below.\\ 
\textbf{Moving Object Detection: } Our estimator takes in measurements by an Inertial Measurement Unit (IMU) and a rolling buffer of asynchronous events $\mathcal{E}=\{e_i\}_{i=0}^{N-1}$ from a monocular event camera, where each event $e_i=(\mathbf{x}_i, t_i, p_i)$ corresponds to a brightness change measurement at pixel $\mathbf{x}_i$, time $t_i$ and sign $p_i$. 
We then compute the \emph{motion compensated mean timestamp image}, $\mathcal{T}(\mathbf{x})$, using the average of the angular rate measurements $\bar \omega$ by the IMU over the duration of the event buffer. $\mathcal{T}$ is defined as 
\begin{align}
    \mathcal{T}(\mathbf{x}) &= \frac{\sum_i (t_i - t_0)\delta(\mathbf{x}-\mathbf{x}_i')}{\sum_i \delta(\mathbf{x}-\mathbf{x}_i')}\\ \nonumber\mathbf{x}'_i&=K\left[\mathcal{I}-[\bar \omega ]_\times  (t_i-t_0)\right]K^{-1}\mathbf{x}_i
\end{align}
Where the $\mathbf{x}'_i$ represent motion compensated coordinates, with calibration matrix $K$. For events that are well explained by a pure rotation, such as when events are triggered in the background, $\mathcal{T}$ will be small, while for independently moving objects, $\mathcal{T}$ will feature higher values. By thresholding $\mathcal{T}$ we can thus segment out independently moving objects in the image plane resulting in the binary map: 
\begin{equation}
    B(\mathbf{x})=\begin{cases}
    1 \quad \text{ for } \rho(\mathbf{x}) > \theta_0 + \theta_1 \Vert \bar \omega \Vert\\
    0 \quad \text{ otherwise }.
    \end{cases}
\end{equation}
Where $\rho(\mathbf{x}) = \frac{\mathcal{T}(\mathbf{x}) - \bar{\mathcal{T}}}{\Delta T}$ is a suitable normalization of the mean timestamp image by its mean, and time window.
We use an adaptive threshold, which depends on the angular rate since higher angular rates will cause a higher average $\rho$.
Having identified pixels of moving objects, we fit maximal rectangles around them by employing the DBSCAN clustering algorithm \cite{Ester96kdd}. DBSCAN works by minimizing a dissimilarity score between pixels assigned to the same cluster. We use the same formulation as in \cite{Falanga20scirob} which minimizes
\begin{align}
    C(\mathbf{x},\mathbf{y}) &=& w_p \Vert\mathbf{x}-\mathbf{y}\Vert + w_v\Vert \mathbf{v}(\mathbf{x})-\mathbf{v}(\mathbf{y})\Vert\\
    \nonumber && + w_\rho\vert\rho(\mathbf{x}) - \rho(\mathbf{y})\vert
\end{align}
where $\mathbf{x}$ and $\mathbf{y}$ are pairs of pixels. Similar to \cite{Falanga20scirob} we also use optical flow similarity as a criterion. We compute it by using dense Lucas Kanade optical flow on consecutive mean timestamp images $\mathcal{T}$. The resulting output of DBSCAN is a set of clusters that we circumscribe with bounding boxes before going to the trajectory estimation step. \\
\textbf{Trajectory Estimation}
Our next step consists of finding the impact point of the object in the plane perpendicular to the robot, which is used by the planner to execute the catching maneuver. To do this we continuously estimate the ballistic trajectory of the object, from noisy 3D position measurements during observation.  While the bounding boxes described above provide us with direction estimates for each point along the parabola, the depth is unobserved for monocular setups. We, therefore, assume a known object size with which we can calculate the object depth according to 
\begin{equation}
    Z = f\frac{W_{\text{metric}}}{W},
\end{equation}
where $W_{\text{metric}}$ and $W$ are the width of the object in the real world and image plane respectively, and $f$ is the focal length of the camera. As a result, we can recover the 3D positions $_\mathcal{C}\mathbf{p}_i$ of the objects in the camera frame $\mathcal{C}$ by back-projecting the ray of the center of the bounding box by a depth $Z$, and transform it to the fixed world frame $\mathcal{W}$ with a known transformation from the robot's odometry. We next fit a parabola of the form
\begin{equation}
    \label{eq:ballistic}
    _\mathcal{W}\mathbf{p}(t) = {_\mathcal{W}\mathbf{p}_0} + {_\mathcal{W}\mathbf{v}_0} t + \frac{1}{2}{_\mathcal{W}\mathbf{g}}t^2
\end{equation}
through these 3D points. In the world frame,  gravity is $_\mathcal{W}\mathbf{g}=[0,0,-9.81ms^{-2}]^T$.
Since these points contain outliers, we use RANSAC to remove them. We successively select random pairs of  3D points, for which we find a solution for $_\mathcal{W}\mathbf{v}_0$ and $_\mathcal{W}\mathbf{p}_0$. For each solution we construct the parabola and count the 3D points within a weighted euclidean distance of the parabola, i.e. inliers, satisfying
\begin{align}
    i\in \text{inliers} \iff \Vert {_\mathcal{W}\mathbf{p}_i} - {_\mathcal{W}\mathbf{p}}(t_i)\Vert_\Lambda\leq \theta\\
    \text{with } \Lambda=R_{\mathcal{W}\mathcal{C}}\text{diag}\left\{\sigma_{xy}^2, \sigma_{xy}^2, \sigma_z^2\right\}^{-1}R_{\mathcal{W}\mathcal{C}}^T 
\end{align}
where $\theta$ is the inlier threshold, $R_{\mathcal{W}\mathcal{C}}$ rotates vectors from camera to world frame, and $\sigma^2_{x/y/z}$ correspond to variances in $x$, $y$, and $z$ direction respectively. Since our 3D object detector exhibit higher noise in the camera $z$-direction, we select $\sigma_x^2=\sigma_y^2=1$ and $\sigma^2_z=5$. 
We select the minimal solution that yields the highest number of inliers, and then fit a $\mathbf{p}_0$ and $\mathbf{v}_0$ which minimizes the square euclidean distance to the inlier points. 
\begin{equation}
    {_\mathcal{W}\mathbf{p}_0}, {_\mathcal{W}\mathbf{v}_0} = \arg \min_{{_\mathcal{W}\mathbf{p}_0},{_\mathcal{W}\mathbf{v}_0}} \sum_{j\in \text{inliers}} \Vert {_\mathcal{W}\mathbf{p}_j} - {_\mathcal{W}\mathbf{p}(t_j)} \Vert^2
\end{equation}
which has a closed-form solution. 

\noindent\textbf{Impact Point Estimation} 
The backend controller expects an estimate of the target position of the net which can catch the object. To estimate this position, we calculate the intersecting point of the parabola trajectory and a plane $\mathcal{P}$ which is spanned in the gravity vector and the y-axis of the robot base. 
% impact point of the object on a impact plane $\mathcal{P}$, expressed in its base frame $\mathcal{B}$. 
The base frame $\mathcal{B}$ is centered on the body of the robot and visualized in Fig. \ref{fig:approach_overview}. We choose $\mathcal{P}$ to share the origin of $\mathcal{B}$. 
% while being spanned by the gravity vector and the y-axis of $\mathcal{B}$.
Points $\mathbf{p}$ on this plane satisfy the relation 
\begin{equation}
    _\mathcal{B}\mathbf{n}_{\text{imp}}\cdot {_\mathcal{B}\mathbf{p}} = 0
\end{equation}
with $_\mathcal{B}\mathbf{n}_{\text{imp}}\doteq {_\mathcal{B}e_y} \times {_\mathcal{B}\mathbf{g}}$
Substituting the parabola equation~\eqref{eq:ballistic}, expressed in body frame $\mathcal{B}$ and solving for $t$, we can find the time to impact, and the impact point as:
\begin{figure}[t!]
    \centering
    \vspace{10pt}
    \includegraphics[width=\linewidth]{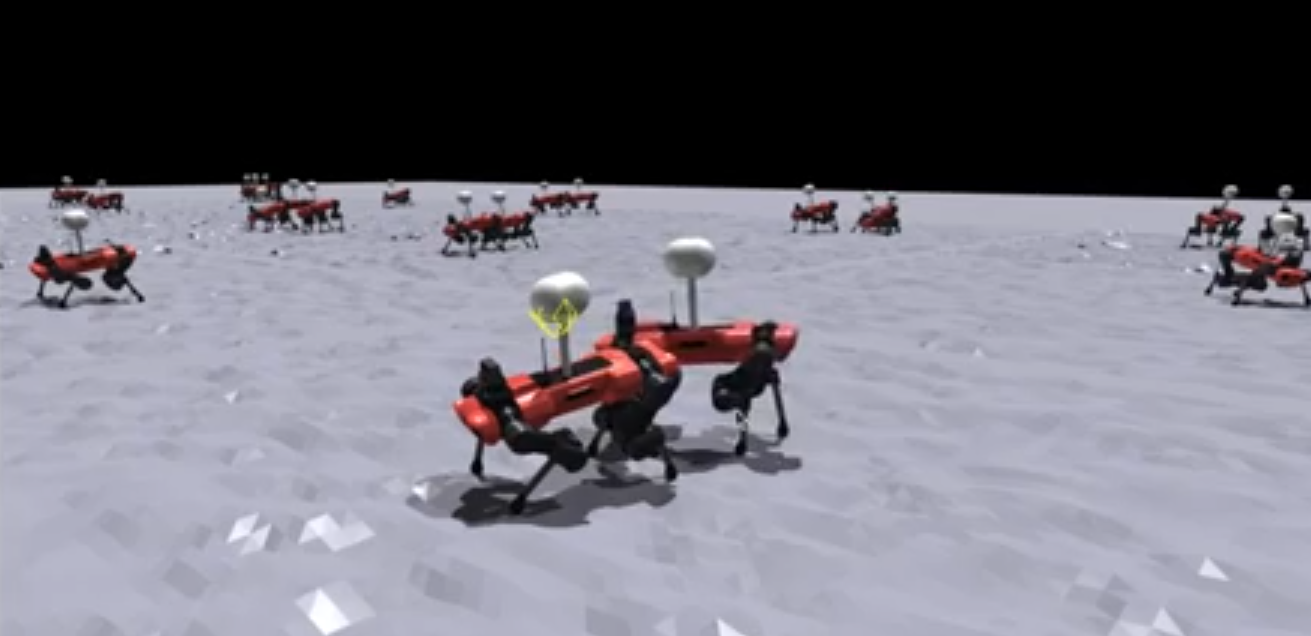}
    \caption{Simulation environment for tracking the impact point $\mathbf{p}_{\text{imp}}$. The agents try to track the yellow spheres with the simulated nets mounted on their base, while avoiding the robot from falling in the presence of random pushes, randomized friction, mass and terrain.}
    \label{fig:sim_env}
    \vspace{-2ex}
\end{figure}
\begin{equation}
_\mathcal{B}\mathbf{p}_\text{imp}={_\mathcal{B}\mathbf{p}}(t_\text{imp}) \text{ with } t_{\text{imp}}=-\frac{_\mathcal{B}\mathbf{n}_{\text{imp}}\cdot {_\mathcal{B}\mathbf{p}_0} }{_\mathcal{B}\mathbf{n}_{\text{imp}}\cdot {_\mathcal{B}\mathbf{v}_0}} 
\end{equation}
%We define the plane $\mathcal{P}$ to be gravity aligned. To arrive to a point that, we pick this point to be on a plane that is perpendicular to the ground and the nominal direction of the throw, and intersects the origin of the robot's base coordinate frame, $\mathbf{o}_{\mathcal{B}}$.
% To find it we solve Eq. \eqref{eq:ballistic} for $p_{0,x}=0$, resulting in a closed form solution for the time to impact, and the resulting impact point $\mathbf{p}_\text{imp}$ on the plane:
% \begin{align}
    % \mathbf{p}_\text{imp} = \mathbf{p}(t_\text{imp}) \quad\text{ with }t_\text{imp} = \frac{-p_{0,z}}{v_{0,z}}.
% \end{align}
To further improve robustness, we apply a median filter to the series of impact point detections, until the robot initiates the catching maneuver.

%% TODO: RANSAC VISUALIZATION HERE?

\subsection{Robot Locomotion Backend}
For object catching, we train a policy that controls each joint of a quadrupedal robot to reach $\mathbf{p}_\text{imp}$ with a net mounted on the robot as fast as possible. We model this catching problem as an \ac{MDP} and use reinforcement learning to solve it in order to achieve agile behavior with low calculation latency. The objective of reinforcement learning is to maximize a discounted sum of rewards over a finite time horizon, \emph{i.e.} finding
\begin{align}
\pi^* = \arg\max_\pi \sum_{t=0}^{T}\gamma^t \mathcal{R}_t 
% \mathbf{s}_{t+1} = f(\mathbf{s}_{t}, \pi(\mathbf{s}_{t})) 
\end{align}
with discount factor $\gamma < 1$, termination time $T$, and reward function $\mathcal{R}_t$. 
% Note that this reward function depends implicitly on $\pi$ via the state transition function $f$. 
% We train the policy by leveraging the highly parallelized implementation \cite{Rudin22corl} of an off-the-shelf on-policy reinforcement learning algorithm, Proximal Policy Optimization (PPO) \cite{Schulman17arxiv}.\\
We used \ac{PPO}~\cite{Schulman17arxiv} to train the policy.

%in which the robot is given the goal position of the net. %Since the application requires us to push the limits of the platform in terms of dexterity and robustness at high speeds, we use large-scale massively parallel deep reinforcement learning~\cite{Rudin22corl}, as it has shown capability exceeding model-based approaches, even enabling robust locomotion in the wild \cite{Miki22sciRob}.

\begin{figure*}[t]
\centering
\vspace{10pt}
\begin{tabular}{ccc}
    \includegraphics[height=4cm]{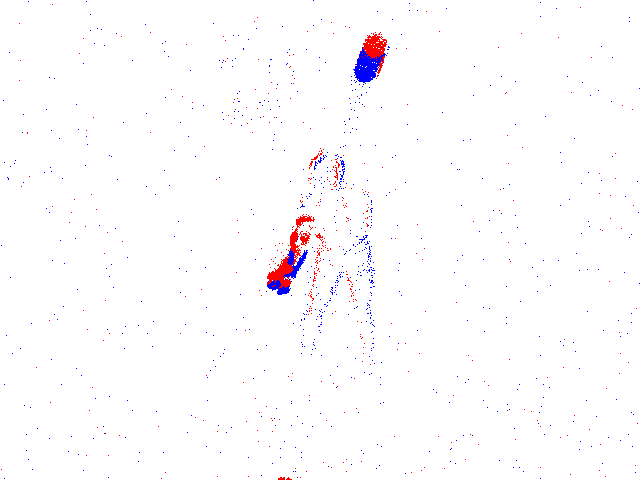}&   
    \includegraphics[height=4cm]{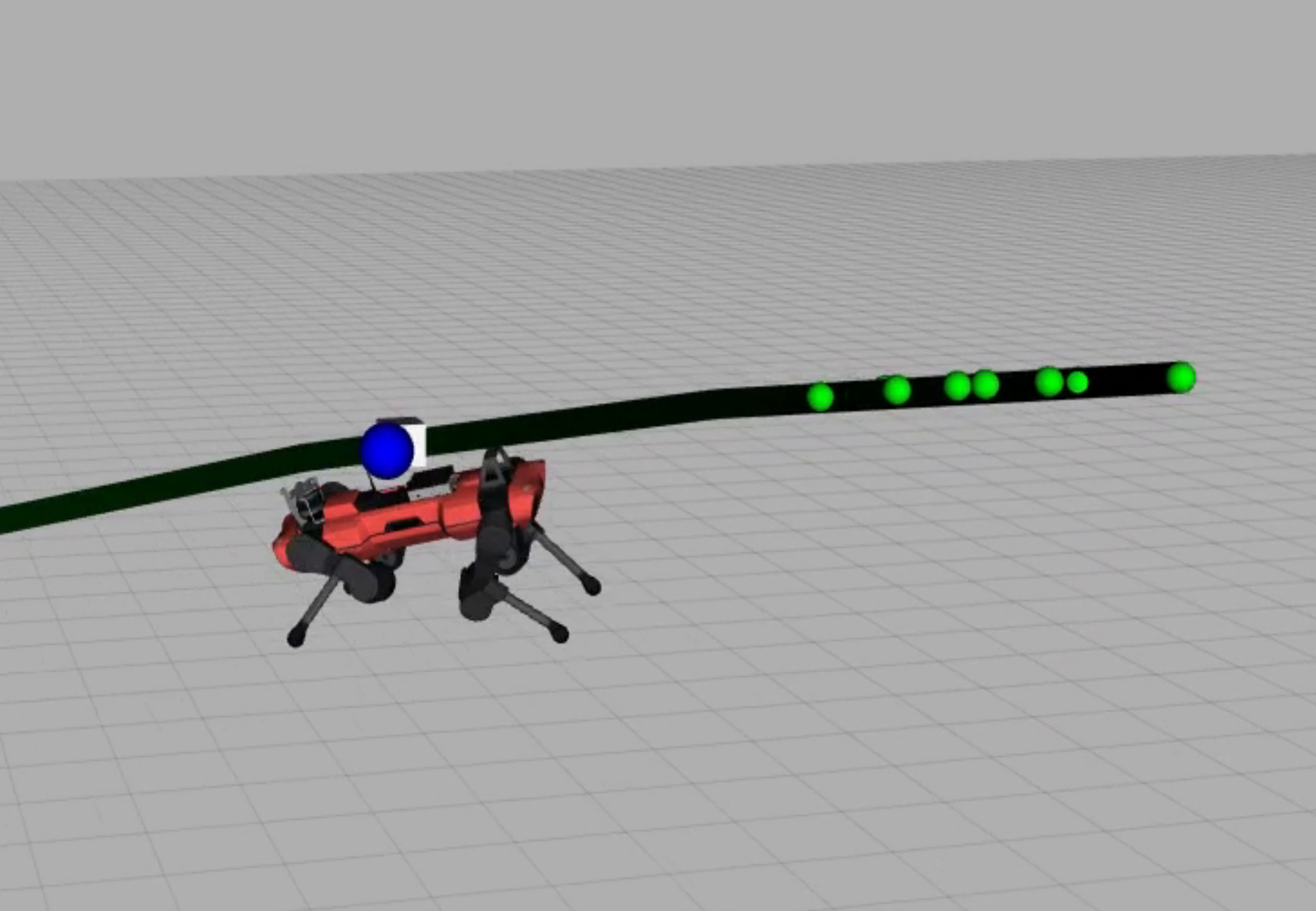}&
    \includegraphics[height=4cm]{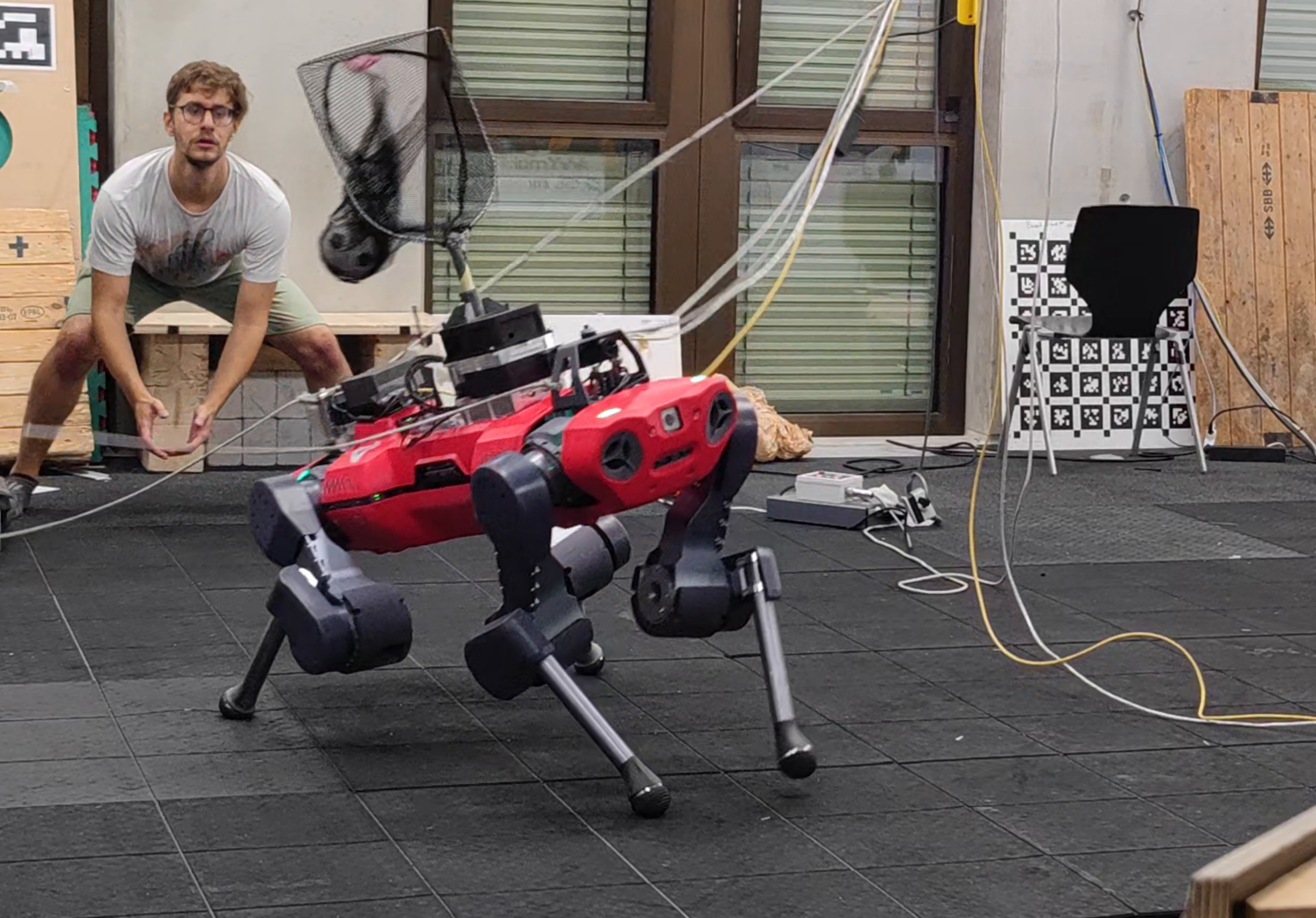}\\
    (a) observed events&
    (b) estimated parabola&
    (c) catching maneuver
\end{tabular}
    \caption{Real world experiments with ANYmal. The DVXplorer event camera (a) observes incoming objects. Our moving object detector produces object detections in 3D, to which we fit a parabola (dark green, b). We then use the recovered impact point (blue circle, b), 
    to execute the catching maneuver (c).} 
    \label{fig:experimental_setup}
    \vspace{-2ex}
\end{figure*}

\noindent\textbf{States, Action space and Policy}
We use ANYmal C~\cite{Hutter2016anymal} as a quadrupedal robot model and use the actuator net~\cite{Hwangbo2019-jk} which simulates the dynamics of the actuator.
% We assume a quadrupedal robot with twelve elastic actuators, three for each leg. 
A policy $\pi$ outputs joint level action $\boldsymbol{\mathbf{\varphi}}_t^*$ that is the differences between the desired and nominal joint position. 
% These outputs are translated to torque commands by a neural network.
At each discrete time step $t$ it takes as input the state $\mathbf{s}_t\in\mathbb{R}^{48}$ defined as 
\begin{equation}
    s_t=[_\mathcal{B}\mathbf{p}_\text{imp} \quad \mathbf{s}^p_t].
\end{equation}
It comprises the impact point expressed in the body frame of the robot, and proprioceptive states $\mathbf{s}^p_t\in\mathbb{R}^{45}$ defined as
\begin{equation}
\mathbf{s}^p_t=[_\mathcal{B}\mathbf{g}\quad _\mathcal{B}\mathbf{v}\quad _\mathcal{B}\boldsymbol{\mathbf{\omega}}\quad \boldsymbol{\mathbf{\varphi}}\quad \boldsymbol{\mathbf{\dot\varphi}}\quad \boldsymbol{\mathbf{\varphi}}^*_{t-1}  ],
\end{equation}
where $_\mathcal{B}\mathbf{g}$ is the gravity expressed in the base frame, $_\mathcal{B}\mathbf{v}$ and $_\mathcal{B}\boldsymbol{\mathbf{\omega}}$ represents body linear velocity and angular velocity,
$\boldsymbol{\mathbf{\varphi}}$ and $\boldsymbol{\mathbf{\dot\varphi}}$ are the joint position and velocity respectively, and $\boldsymbol{\mathbf{\varphi}}^*_{t-1}$ is the previous action.
% Due to the high agility and robustness required for ball catching, we opt for a data-driven policy, as was successfully deployed in works such as \cite{Miki22sciRob,Song21iros}. We model our policy as a five-layer multilayer perceptron (MLP) with ELU activations, which maps $\mathbf{s}_t$ to $\boldsymbol{\mathbf{\varphi}}^*_t$.    \\

\noindent\textbf{Environment setup}
We use Isaac gym~\cite{Makoviychuk2021-th} to build the simulation environment and leverage the highly parallelized implementation on a \ac{GPU}~\cite{Rudin22corl}.
In the simulation environment, we spawn a quadrupedal robot equipped with a net on terrain and randomly sample target net position (Fig. \ref{fig:sim_env}).
% The policy tries to make the center of the net as close as possible to the target net position.
The target net position is randomly sampled around the current net position within the range of $\pm30$ cm on the x-axis, $\pm 80$ cm on the y-axis, and $\pm 15$ cm on the z-axis in the robot base frame.
We resample the target net position every 1.0 seconds with a probability of 0.2. To make the policy robust to noisy sensor readings and unexpected foot trapping, we add noise to the observation and a slight roughness to the terrain as seen in Fig. \ref{fig:sim_env}.
The maximum episode length is 20 sec and we terminate the episode when the robot base or thigh hits the ground.

\noindent\textbf{Rewards}
To make the policy track the target net position, we used the following rewards. \\
% with a smooth motion without falling, we used the following rewards. \\
\indent Target tracking : $r_\text{tr} = \exp(-10\|_\mathcal{B}\mathbf{p}_\text{imp} - _\mathcal{B}\mathbf{p}_\text{net}\|^2)$ \\
In addition, we have the following penalty rewards to make the motion smooth. \\
\indent Base linear velocity: $r_{lv} = v_z^2$ \\
\indent Base angular velocity: $r_{av} = \omega_x^2 + \omega_y^2$ \\
\indent Joint velocity: $r_{v} = \|\boldsymbol{\mathbf{\dot\varphi}}\|^2$ \\
\indent Joint acceleration: $r_{a} = \|\boldsymbol{\mathbf{\ddot\varphi}}\|^2$ \\
\indent Joint torque: $r_{t} = \|\boldsymbol{\mathbf{\tau}}\|^2$ \\
\indent Action difference: $r_{ac} = \|\boldsymbol{\mathbf{\varphi}}^*_{t-1} - \boldsymbol{\mathbf{\varphi}}^*_{t}\|^2$ \\
\indent Base orientation: $r_{o} = \|_\mathcal{B}\mathbf{g}_{xy}\|^2$ \\
The final reward is defined as follows by checking the behavior of the policy.
\begin{eqnarray}
    r &=& 4r_\text{tr} - 0.1r_{lv} - 0.05r_{av} - 3\cdot 10^{-6} r_{v}  -5\cdot 10^{-7} r_{a} \nonumber \\
    &&- 2\cdot 10^{-5} r_{t} - 0.03 r_{ac} - 0.2r_{o} 
\end{eqnarray}

\begin{figure}[t!]
    \centering
    \includegraphics[width=\linewidth]{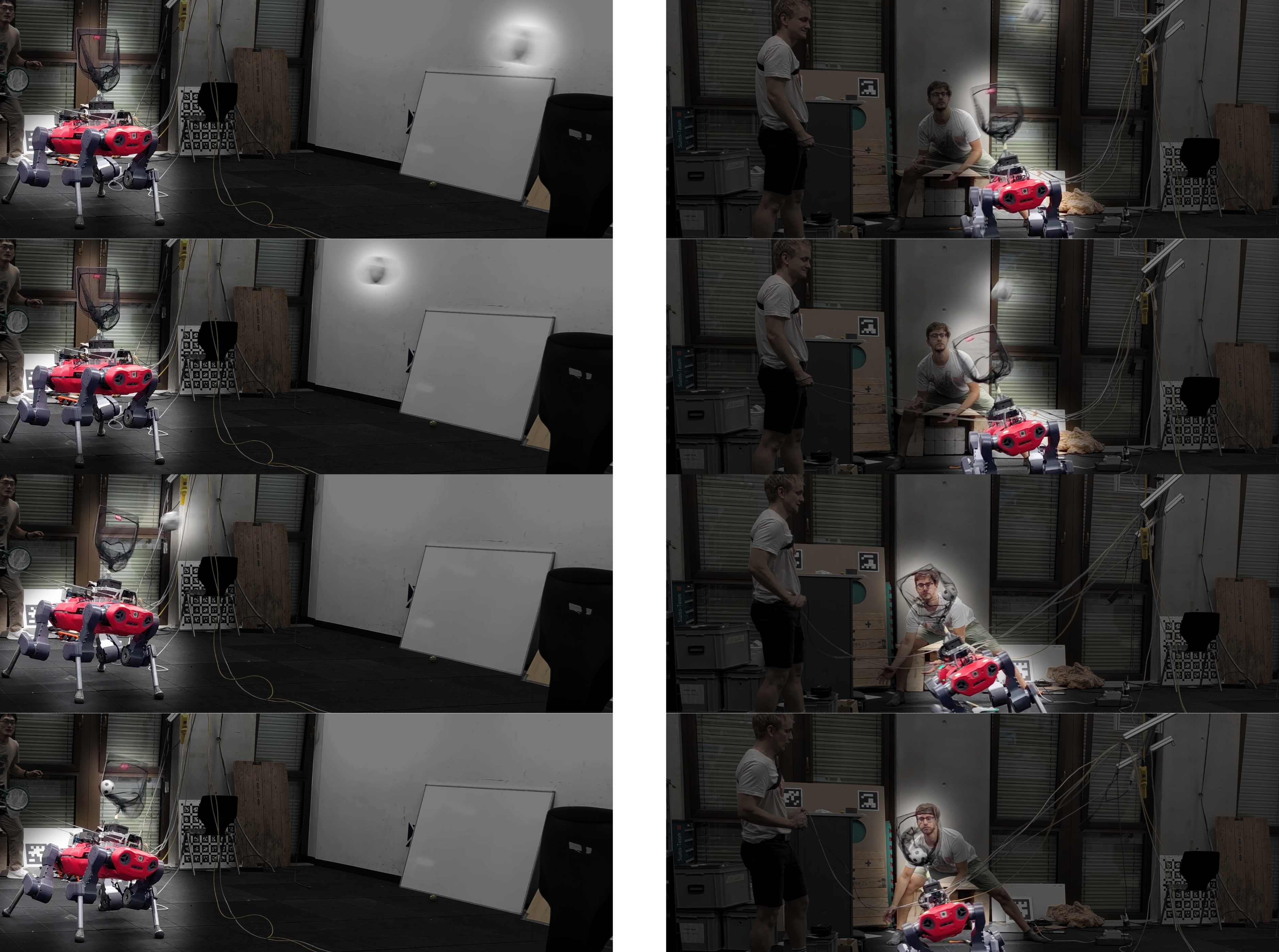}
    \caption{Image sequence illustrating two successful catches by our method. As can be seen, the robot is required to perform highly agile tilting and side walking behaviors to catch the ball. The maximum ball speeds in both columns are 11.5 m/s and 10 m/s respectively.}
    \vspace{-2ex}
    \label{fig:agil_maneuver}
\end{figure}

\iffalse 
\begin{itemize}
    \item Perception frontend for 3D object position and velocity estimation
    \item Robot backend policy perception and proprioception to action.
\end{itemize}
\subsection{Perception frontend}
\begin{itemize}
    \item Event camera data
    \item Motion compensation and Thresholding
    \item Network architecture
\end{itemize}
\subsection{Robot backend}
\begin{itemize}
    \item Describe PPO \cite{schulman2017proximal} as Taka did in \cite{miki2022learning}
\end{itemize}
\subsection{Training details}
\begin{itemize}
    \item Simulator details
    \item First stage training: supervised perception network
    \item Teacher policy training. 
    \item End-to-end training of perception and action student.
\end{itemize}
\fi

%\section{LEARNING TO CATCH FAST OBJECTS}
%\input{chapters/learning_to_catch}

%\section{DETECTION OF DYNAMIC OBJECTS}
%\input{chapters/vision}

\section{Experiments}
% , with deviations from the net center in the range of $Z [cm]$ to $W [cm]$. Average ball speeds and deviations from the net center were tracked with two separate cameras.
\iffalse
\textbf{Simulated Experiments} To test our policies against ground truth data in a highly configurable setting, we set up an Isaac Gym environment coupled with event camera simulation that allows running the trained policy $\pi(a_t | o_t)$ and the vision frontend in a wide range of domain randomized environments. Similarly to our later hardware experiments, we fire a ball with a randomized texture at the robot from $5 [m]$ with a speed in the range of $8-20 [m/s]$ with deviations from the net centre in the range of $0-1 [m]$. To generate background clutter, the robot is surrounded by four walls and a floor that all have randomized textures. The scene is illuminated by white ambient lighting of constant intensity, assuming a classical Phong reflection model. The mean error of the detected $\mathbf{p_{imp}}$ and the success rate plotted against the ball speed and $\mathbf{p_{imp}}$'s deviation from the net centre can be seen on \ref{fig:method}. This simulation is not only useful for benchmarking our current approach, but can pave the way for end-to-end learning of event-to-maneuver policies as well later.\\
\fi
\noindent\textbf{Hardware setup} We directly deploy the policy learnt in simulation on the real robot, finding that it required no additional fine-tuning to run in the real world.  We use the  ANYmal C quadruped robot \cite{Hutter2016anymal}, coupled with a VGA resolution event camera, an IniVation DVXplorer~\cite{Lichtsteiner08ssc}, fitted with a $90^{\circ}$ field-of-view (FoV) objective. For catching, a landing net with a diameter of $0.3$ m was mounted on the quadruped with its center being $0.5$ m above the base frame. The detection algorithm runs on an NVIDIA Jetson Orin developer kit and 
% a state-of-the-art edge AI platform with a 12-core ARM Cortex CPU and 2048 NVIDIA CUDA cores. While our current setup is not running anything in CUDA.
the catching policy was deployed on the robot's onboard PC, an Intel core i7 8850H with 6 cores and 12 threads. To display the debugging visualization, two Ethernet cables were attached to the laptop PC, but all compute necessary to execute the catching maneuver is running on board. With this hardware setup, our vision algorithm is capable of running at up to 100 Hz.\\

%\textbf{Frame-based baseline} Our robot of choice, ANYmal C, has a 1.5 MP RGB FLIR Blackfly BFS-GE-16S2C-BD2 camera with a wide angle $110^{\circ}$ FoV objective that is capable of outputting frames at $75[Hz]$. Usin
We test the catching pipeline by throwing balls manually in a cluttered environment. To test texture generalization, two different rubber balls were used, each with a diameter of $10$cm. The balls were thrown manually at average speeds between $5-15$m/s, with net center deviations in the range of $~0.6$ m. At these speeds, the incidence angle of the ball was around $90\pm 15^\circ$ with respect to the net. A preview of the experiments, including event camera and parabola visualization can be viewed in Fig. \ref{fig:experimental_setup}

\noindent\textbf{Evaluation Proceedure} We validate our method by measuring two forms of success rate, one for the vision algorithm, and one to evaluate the overall performance of the system. In the first, we measure the percentage of impact point measurements that have an error lower than the net radius and report the results of this experiment in Tab. \ref{tab:results_vision}. To evaluate the overall performance we report the percentage of successful catches within a working range of 0.6 m from the initial net position and study the success rate for different throw speeds and deviations from the initial net center. We selected this working range, since beyond it, ANYmal needs to perform significant sideways locomotion, instead of lunging, which slows down the maneuver. These results are summarized in Tab. \ref{tab:results_robot}. We collect data for a total of 22 throws, for which the ball enters the feasible range of 0.6 m from the robot. While estimated impact points were measured in our visualization (see Fig. \ref{fig:experimental_setup}), measurements of ball speeds and impact points in the real world were measured using footage from a 60 fps external camera.

\noindent\textbf{Results}
We find from Tab. \ref{tab:results_vision} that our vision algorithm only degrades at speeds $>9$ m/s and deviations from the net position of $>0.4$m. At 15 m/s and 0.6 m deviation, we still capture 50.25\% of the impact points within a reasonable accuracy for catching. The degradation can be explained by two factors: as speed increases, fewer samples are collected to fit a parabola, reducing its overall accuracy. Similarly, throwing the ball further out, means that the ball quickly leaves the FoV, resulting in fewer observations. In Tab. \ref{tab:results_robot} we find that catching success rate remains above 80\% reaching a maximum at $12$ m/s. Two examples of successful catching maneuvers can be found in Fig. \ref{fig:agil_maneuver}. We found that too high speeds cause vision to fail more often due to too few samples being recorded. Additionally, the robot cannot react fast enough to catch the objects.
Conversely, at too low speeds, our vision algorithm can correctly identify where the robot should go, but the low speeds cause higher variability in the range of the trajectory and the impact angle at which the ball trajectory reaches the robot becomes steeper. 
If objects arrive at steep impact angles, the effective area with which the robot can catch the object is reduced. The impact angle is not directly observed by our policy, which reduces its robustness to this factor.

% As the angle of the net is almost vertical, this may reduce the tolerance of net position. 
% A too short or long ranges may force the robot to move in the x direction, which is inherently slower than simple sideways tilting, reducing the success rate.

\begin{table}[]
\centering
\vspace{10pt}
\begin{tabular}{cccc}
\hline
\multicolumn{4}{c}{\textbf{Object Speed {[}m/s{]}}} \\  
\textless 8.0 & \textless 10.0  &\textless 12.0 &  \textless 15.0  \\ \hline
$81 \%$ & $88 \%$  &  $92 \%$ & $ 83 \%$   \\ \hline
\end{tabular}
\caption{Success rate of ball catching for different object speeds.}
\label{tab:results_robot}
\end{table}

\begin{table}[]
\centering
\begin{tabular}{l|cc}
\hline
\multicolumn{1}{c|}{\multirow{2}{*}{\textbf{\begin{tabular}[c]{@{}c@{}}Distance from \\  Net Position {[}m{]}\end{tabular}}}} & \multicolumn{2}{c}{\textbf{Object Speed {[}m/s{]}}} \\  
\multicolumn{1}{c|}{}                                                                                                         & \textless 9    & \textless 15  \\ \hline
\textless 0.4 &     $93,33  \%$     &  $88,88 \%$   \\
\textless 0.6 &     $73,68 \%$       & $50,25 \% $      \\ \hline
\end{tabular}
\caption{Success rate of our vision algorithm. Successful impact point detections are within a net radius from the ground truth position.}
\label{tab:results_vision}
\end{table}

%\begin{table}[]
%\centering
%\begin{tabular}{ccc}
%\hline
%multicolumn{3}{c}{\textbf{Object Speed {[}m/s{]}}} \\ \hline
%textless 5    & \textless 10.0   & \textless 15.0   \\ \hline
%             $100\%$  &   $100 \%$              &     $63 \%$         \\ \hline
%\end{tabular}
%caption{Success rate of our vision algorithm. Successful impact point detections are %within a net radius from the ground truth position.}
%\label{tab:results_vision}
%end{table}

\iffalse
\subsection{Hardware Setup}
\textbf{}
\begin{itemize}
    \item Describe hardware setup for testing
\end{itemize}
\subsection{Baselines}
\begin{itemize}
    \item Falanga detector + simple waypoint following
    \item Frame-based detector + simple waypoint following
\end{itemize}
\subsection{Results}
\begin{itemize}
\item Plot of success rate as a function of object speed (how will we measure this?)
\item Comparison with image-based and model-based approaches in terms of latency, success rate and maximum speed where still above 90\%.
\end{itemize}
\fi 

\section{Discussion}
\iffalse
\begin{itemize}
\item Limitations of event cams: low resolution (small FoV) and higher noise.
\end{itemize}
\fi
The current work demonstrates agile catching of high-speed objects between $5-15$ m/s with a quadruped robot using an event camera for estimating the ballistic trajectories of the flying objects. Being the first example of catching with a quadruped robot using onboard sensing, our pipeline demonstrates agile catching skills and catches balls coming at a $0.4 $ m deviation from the current center of the landing net. Importantly, as shown in \ref{tab:results_vision}, the onboard event-based ball-tracking pipeline performs reliably on a larger range of $0.6$m, making sure that the estimations provided to the rather dynamic motion policy will always be adequate. While this approach was enough for the first working demo of quadruped catching with onboard sensing, improvements could be made to improve the robot's "range" to the side by for example using a higher FoV camera. Since our method is monocular, it requires knowledge of the object size. However, this requirement could be easily lifted by using a stereo setup, which would also reduce depth uncertainty making us less reliant on filtering. 
Also, the current control strategy was simply reaching the target position as quickly as possible. By taking into account the ball movement, sensor property and detection pipeline during the training, the trained policy could further improve its capability. Finally, to eliminate failure cases due to steep impact trajectories, which reduce the effective net area, we may consider augmenting our policy input by the velocity vector at the impact point.

\section{Conclusion}
On the quest toward animal-like agility and robustness, current quadruped robots have reached a high degree of versatility through the use of both proprioceptive and exteroceptive sensors. However, commonly used sensors such as RGB cameras or LiDARs currently suffer from major limitations due to the bandwidth-latency trade-off, meaning that both low-latency and low-bandwidth cannot be achieved with current systems. In this work, we break this cycle by outfitting a quadrupedal robot with an event camera, which does not have this tradeoff. We design a method that combines a high-speed visual frontend with a learning-based planner to perform the complex task of agile ball-catching. Thanks to the low latency of the event camera, as well as the robustness of the planner we reach an 83\% success rate while catching objects thrown at up to 15 m/s.  


%%%%%%%%%%%%%%%%%%%%%%%%%%%%%%%%%%%%%%%%%%%%%%%%%%%%%%%%%%%%%%%%%%%%%%%%%%%%%%%%

%%%%%%%%%%%%%%%%%%%%%%%%%%%%%%%%%%%%%%%%%%%%%%%%%%%%%%%%%%%%%%%%%%%%%%%%%%%%%%%%

%%%%%%%%%%%%%%%%%%%%%%%%%%%%%%%%%%%%%%%%%%%%%%%%%%%%%%%%%%%%%%%%%%%%%%%%%%%%%%%%
\iffalse
\section*{APPENDIX}

\begin{itemize}
    \item Furter resources on mono, stereo and event camera latency.
    \item Datasheet extracts to back our claims, if needed.
    \item Elaboration of the testing setup.
\end{itemize}
\fi
%\section*{ACKNOWLEDGMENT}
%Funding and nothing else, unless there was another researcher who gave alot of ideas.\davide{Funding is already listed on page one. Remove this text if no researcher is to be acknowledged.}
%Thank RSL for robot hardware & integration help, thank RPG for camera. Necessary?

%%%%%%%%%%%%%%%%%%%%%%%%%%%%%%%%%%%%%%%%%%%%%%%%%%%%%%%%%%%%%%%%%%%%%%%%%%%%%%%%

% Bibliography - max 20! citations
\bibliographystyle{IEEEtran}

\bibliography{references}

\end{document}